\documentclass[conference]{IEEEtran}
\IEEEoverridecommandlockouts
\usepackage{cite}
\usepackage{amsmath,amssymb,amsfonts}
\usepackage{algorithmic}
\usepackage{graphicx}
\usepackage{textcomp}
\usepackage{xcolor}

\usepackage{flushend}

\newcommand{\citet}{\cite}
\newcommand{\citep}{\cite}

\title{Semi-supervised Text Regression with Conditional Generative Adversarial Networks}

\author{\IEEEauthorblockN{\textbf{Tao Li}}
\IEEEauthorblockA{Purdue University \\
\texttt{taoli@purdue.edu}}
\and
\IEEEauthorblockN{\textbf{Xudong Liu}}
\IEEEauthorblockA{ObEN, Inc. \\
\texttt{xudong@oben.com}}
\and
\IEEEauthorblockN{\textbf{Shihan Su}}
\IEEEauthorblockA{California Institute of Technology \\
\texttt{ssu@caltech.edu}}
}

\begin{document}

\maketitle

\begin{abstract}
Enormous online textual information provides intriguing opportunities for understandings of social and economic semantics. In this paper, we propose a novel text regression model based on a conditional generative adversarial network (GAN), with an attempt to associate textual data and social outcomes in a semi-supervised manner. Besides promising potential of predicting capabilities, our superiorities are twofold: (i) the model works with unbalanced datasets of limited labelled data, which align with real-world scenarios; and (ii) predictions are obtained by an end-to-end framework, without explicitly selecting high-level representations. Finally we point out related datasets for experiments and future research directions.
\end{abstract}

\section{Introduction}\label{sec:intro}
With millions of textual information uploaded every day, the Internet embeds tremendous data of social and economic phenomena, and have attracted consistent interests not only from sociologists and economists but also statisticians and computer scientists. For example, \citet{joshi2010movie} forecasted movie revenues using online reviews; based on social media data, \citet{lampos2010tracking} monitored flu pandemic and \citet{lampos2013user} predicted election results.

To our best knowledge, the concept of text regression was first introduced by \citet{kogan2009predicting} who described it as: given a piece of text, predict a real-world continuous quantity associated with the text's meaning. They applied a linear model to estimate financial risks by using financial reports directly and claimed a significant outperformance compared to previous methods.
Subsequently, several linear text regression models were proposed; to name a few: \citet{volkova2014inferring,lampos2014predicting,preoctiuc2015analysis}.

Although easy for interpretation and implementation, linear models rely heavily on specific selections of high-level textual representations and fail to properly capture complicated distributions. Recent successese of deep neural networks in the field of computer vision (e.g., \citet{ledig2017photo} and \citet{liu2018mining}) encourage reseachers to discover their potential in natural language processing. Unlike image synthesis, using deep networks for natural language generation (NLG) is notoriously difficult \citep{li2018toward}, as the feature space of a sentence is discrete and thereby discontinuous and non-differentiable. \citet{kusner2016gans} attacked this issue by using one-hot vectors obtained from softmax function for backpropergation. \citet{lin2017adversarial} used ranking scores instead of real/fake prediction for the objective function of the discriminator.

\begin{figure*}[ht!]
\begin{center}
\includegraphics[width=1.0\textwidth]{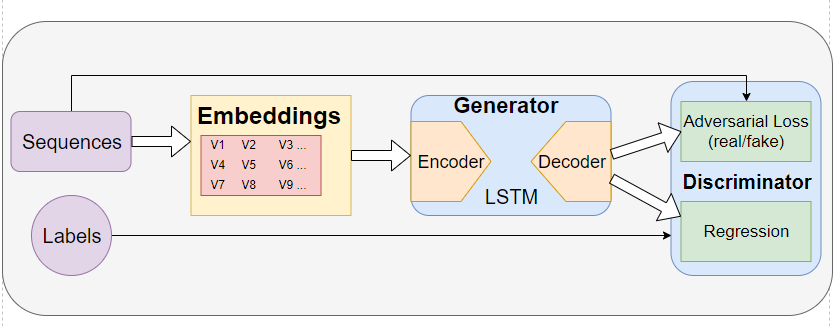}
\end{center}
\caption{Architecture of the TR-GAN model.}
\label{fig:overview}
\end{figure*}

Our idea of using GANs for text regression was inspired by recent advances in NLG (e.g., \citet{yu2017seqgan} and \citet{li2017adversarial}). We further shift the focus from realistic language synthesis to the generation of adversarial samples
from a LSTM \citep{hochreiter1997long}, who competes against a discriminator for regression (see Figure \ref{fig:overview}).
The performance of our model is guaranteed by deep neural networks' power of capturing complicated distributions especially when obtained in an adversarial manner.
The capability of training with limited supervision also facilitates promising future applications.


The rest of the paper is organized as follows:
in Section \ref{sec:literature} we discuss existing text regression techniques and previous works in semi-supervised learning with GANs;
the model is detailed in Section \ref{sec:model};
we conclude the paper in Section \ref{sec:conclusion} by future works.

\section{Related Work}\label{sec:literature}

\subsection{Text Regression}

Previous attempts at text regression mainly focused on linear models.
\citet{kogan2009predicting} adopted a support vector regression (SVR) \citep{drucker1997support} in financial reports to predict the volatility of stock returns, a widely used measure of financial risk, and reported a significant outperformance compared to state-of-the-arts.
To correlate movies' online reviews and corresponding revenues, \citet{joshi2010movie} extracted high-level features of textual reviews and incorporated them into a elastic net model \citep{zou2005regularization}.
\citet{lampos2013user} exploited a multi-task learning scheme that leverages textual data with user profiles for voting intention prediction.
As mentioned earlier, linear models sometimes are oversimplified and fail to properly capture real-world scenarios.
\citet{bitvai2015non} proposed the first non-linear model, a deep convolutional neural network, for text regression which surpassed previous state of the art even with limited supervision.



\subsection{Semi-supervised Learning}
Semi-supervised learning tackles the problem of learning a mapping between data and labels when only a small subset of labels are available. Earlier approaches of generative models with semi-supervised learning consider Gaussian mixture models \cite{zhu2006semi} and non-parametric density models \cite{kemp2004semi}, but suffer from limitations of scalability and inference accuracy. Recently \citet{kingma2014semi} addresses this problem by developing stochastic variational inference algorithms for join optimization of model and variational parameters.

Since generative adversarial networks (GANs) has been shown to be promising in generating realistic images \cite{goodfellow2014generative}, several approaches have been proposed to use GANs in semi-supervised learning. \citet{springenberg2015unsupervised} extends the discriminator ($D$) to be a $K$ class classifier with objective function to minimize prediction certainty on generated images, while generator aims for maximize the same objective. \citet{odena2016semi} augments the $K$ class discriminator to include a $K + 1$ label as fake for the generated images. These work have shown that incorporating adversarial objectives can make the learning of classifier robust and data efficient. While previous works mainly focus on classification setting, in our work, we extend the GAN based semi-supervised learning to regression task.

\section{The TR-GAN Model}\label{sec:model}
In this section, we detail the conditional generative adversarial network for text regression in a semi-supervised setting (TR-GAN). We first introduce the word embedding method.

\subsection{Word Embedding}
Word embedding method learns a high dimension representation for each word, thereby incorporate semantic information that cannot be captured by the single token. In our work, we adopted pretrained word embedding for each word in the text input. Then each document in data can be represented by a $ D \times N$ matrix, where $D$ is the number of words in the document and $N$ is the dimension of word embedding in the pretrained model.

\subsection{Model Architecture}

As illustrated in Figure \ref{fig:overview}, the network architecture is a conditional GAN with a generator and a discriminator.
A long short-term memory network (LSTM) \citep{hochreiter1997long} is deployed as the generator for natural languages. As the embedding is fed into LSTM, the generator is a LSTM-based sentence decoder. The discriminator is a convolutional neural network (CNN) \citep{kalchbrenner2014convolutional}, where serval residual blocks \citep{he2016deep} are followed by batch normalization with $ReLU$ as the activate function. Subsequently, two fully connected layers are finalized for adversarial learning and the regression task.

The objective function adopt mean absolute error (MAE) for regression tasks and adversarial loss for sequence generation. Not only can this model generate realistic sentences through the optimized generator but the discriminator is also trained as a regression model for multiple prediction tasks (e.g., auto sales prediction, public opinion tracking, and even epidemiological surveillance from social media), which are of great interest to a wide range of stakeholders.

\section{Future Work}\label{sec:conclusion}
We are excited about the idea of using GANs for text regression. Given the nature of the TR-GAN model, it is not challenging to find an experimental dataset; for example, \citet{li2018youtube} collected 50,000 textual comments below YouTube videos, among which 20,000 are labelled by state-of-the-art algorithms and 1,000 are labelled manually. We also are interested to see how the generated languages look like, given that existing literatures of using GANs for NLG merely report original experimental results but instead numerical metrics.

\section*{Acknowledgments}

We thank Hao Peng and Kantapon Kaewtip for insightful discussions. The idea of this work originally came out during discussions of \cite{wang2018cooperative} and \cite{gong2018cooperative}.

\bibliographystyle{IEEEtran}
\bibliography{db}

\end{document}